# Accurate Gaze Estimation using an Active-gaze Morphable Model

Hao Sun[a,**], Nick Pears[a]

[a]*University of York, YO10 5DD, York, UK*

ABSTRACT

Rather than regressing gaze direction directly from images, we show that adding a 3D shape model can: i) improve gaze estimation accuracy, ii) perform well with lower resolution inputs and iii) provide a richer understanding of the eye--region and its constituent gaze system. Specifically, we use an 'eyes and nose' 3D morphable model (3DMM) to capture the eye-region 3D facial geometry and appearance and we equip this with a geometric vergence model of gaze to give an 'active-gaze 3DMM'. We show that our approach achieves state-of-the-art results on the Eyediap dataset and we present an ablation study. Our method can learn with only the ground truth gaze target point and the camera parameters, without access to the ground truth gaze origin points, thus widening the applicability of our approach compared to other methods.

## 1. Introduction

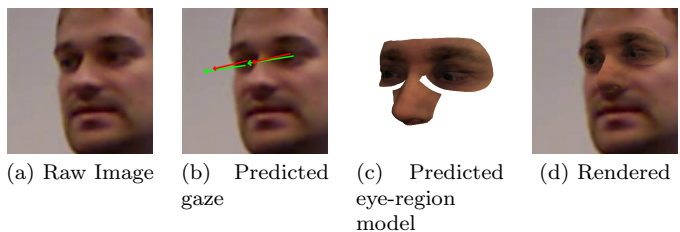

Fig. 1. Active-gaze 3DMM fitting example. (a) Raw input image, (b) predicted gaze directions (red) compared to ground truth (green), (c) predicted eye region model, (d) eye region model rendered and overlaid on raw input image.

Human eye gaze estimation enables the visual understanding of human intention, with high utility in human-computer interaction and virtual reality. Previous works have built an eye-region model [1], and a full head model [2], that can be fitted to given images, which thereby provides a gaze direction estimation. Also, appearance-based methods that regress gaze directions directly from RGB input images using deep networks without the use of 3D shape models have been popular [3]. Compared to these appearance-based methods, model-based methods are less competitive in regard to gaze estimation accuracy, due to a deep neural network's feature extraction and nonlinear fitting ability. However, most appearance-based gaze estimation methods predict only a gaze direction (azimuth-elevation orientation), but no other information about the 3D geometry of the gaze or the eye-region. Current literature has different gaze origin representations (*e.g.* eyeball centres or a point on the face), which requires additional effort to make performance comparisons [4].

We propose an end-to-end method that combines both appearance-based and model-based elements. Our method reconstructs the 3D eye-nose region and thereby avoids the highly-variable mouth-jaw area, so that it can more accurately predict gaze direction. We employ an *eyes-and-nose* 3D morphable model (3DMM) and, crucially, we equip this with a geometric vergence model of gaze. We call this an '*active-gaze* 3DMM'. Specifically, this enables the combined rotation of the eyeballs for the expression of gaze under certain geometric constraints, such as coplanarity of the gaze vectors. As a result, we can model the correlations between the face and the left and right eyeballs. This ensures both accurate gaze estimation and that the eyeball

---

[**]Corresponding author

positions are consistent with both the face geometry and head pose, see Fig. 1.

The core idea of adding a model to the appearance-based method is to provide richer information. For example, we can trivially calculate the subject's inter-ocular distance from the model. Another example is that, when designing wearable devices such as a VR headset or smart glasses, both the eye-region geometry information and the gaze information are important. Under such circumstances, our approach provides both the eye-region geometry and a much more accurate gaze estimation compared to purely model-based methods.

Most image autoencoder 3D reconstruction methods from monocular RGB images focus on faces [5, 6]. Typically, their 3D face models only model the eyeball surface area as part of the face, and the gaze directions are not explicitly modelled. Our method both takes advantage of the image autoencoder architecture *and* models the specific eye-region area, designing gaze information into the model. In summary, our main contributions are: i) An *active-gaze 3DMM* that focuses on the more rigid 'eyes and nose' region and that is equipped with a geometric eye vergence model for regularisation. ii) Demonstration that the active-gaze 3DMM increases gaze estimation accuracy and versatility. iii) Demonstration of the method's adaptability, when only ground truth 3D gaze targets are available, with no access to gaze origin information. To the best of our knowledge, we are the first to propose a gaze estimation method that combines both an appearance-based method and a model-based method.

## 2. Related Work

### 2.1. Face and Eyes 3D Morphable Models (3DMMs)

Face 3DMMs were introduced more than two decades ago by Blanz and Vetter [7] and perhaps is the most widely-employed technique in recent statistical 3D face modelling applications. Such 3DMMs model a linear or non-linear 3D facial space using a latent representation that can be constructed in a number of different ways. Examples include PCA [8, 9, 2], dictionary learning [10], wavelet decomposition [11], Gaussian mixture models [12] and neural nets [6]. Apart from the general face 3DMMs, there are several approaches that bring more focus to eyeball modelling. Bérard *et al.* were the first to build a parametric model of eyeballs. The quality of this eye model is high, but the reconstruction process is semi-automatic. Wood *et al.* [1, 13] attempt to build an eye-region model of the single eye and use model fitting to estimate gaze. Ploumpis *et al.* [2] propose a method for building a complete head morphable model that includes eyeballs. The eye-region modelling is similar to the approach of Wood *et al.* and is blended into the head model.

Amongst the publicly-available 3DMMs, we choose the FLAME [14] model to build our eye-region model since it has both eyeballs and can form a minimal eye-region model for both eyes, see Fig. 1c.

### 2.2. 3D Reconstruction from Monocular RGB

Numerous reconstruction methods have been proposed and they generally fall into three categories: generative, regression and generative-regression hybrid. Generative methods focus on generating a 3D model to fit the target data [15]. The approaches proposed by Wood *et al.* [1] and Ploumpis *et al.* [2] both fall into this category. Regression methods, recently popular due to deep learning advances, focus on regressing the model parameters directly via deep networks [16, 17]. The third category was firstly proposed by Tewari *et al.* [5], and adopted by many other works [18, 6]. This approach usually trains a joint autoencoder model that encodes the model parameters via the regression method, decodes the regressed model parameters, and reconstructs the original input. In contrast to our work, all of the mentioned face autoencoders focus on full face reconstruction and use only a mesh surface to model the eyeball, and the appearance of different gaze directions is not present or modelled via texture.

### 2.3. Appearance-based Gaze Estimation Methods

Appearance-based gaze estimation using a deep neural network to regress gaze directions has been very popular recently. A number of datasets containing RGB images and gaze labels have been published that have enabled rapid progress [19]. Along with the datasets, various appearance-based methods using different input representations, *e.g.*, eye images [20, 21, 22], face images [23, 24] and both [25, 19]. They also use different network architectures (CNNs, attention-based or combined) and different gaze representations (gaze originates from the eyeball and gaze originates from face centre) [4]. Prediction using high-resolution eye images has been the mainstream of this area, but face features were found to provide additional information for gaze estimation [26].

### 2.4. Model-based Gaze Estimation Methods

On the more traditional side, model-based gaze estimation often involves a geometric eye model that is fitted to detected eye features, *e.g.*, corneal reflections, pupil centre, iris contour, and eye landmarks [1, 2]. Some of them require dedicated devices such as RGB-D camera and infrared light. For the RGB camera approaches, Wood *et al.* [1] and Ploumpis *et al.* [2] use eye-region landmarks to fit their 3DMM models.

## 3. Proposed Method

Our architecture in Fig. 2, shows that the raw image $\mathbf{I}$ is fed to the encoder to regress eye-region reconstruction parameters $\mathbf{z}_M$ and eye rotation parameters $\mathbf{z}_E$. The eye-region parameters are defined as follows: $\mathbf{z}_M = (\mathbf{z}_S, \mathbf{z}_A, \mathbf{r}, \mathbf{T}, f)^T$, where $\mathbf{z}_S$ are shape parameters, $\mathbf{z}_A$ are texture parameters, $\mathbf{r}, \mathbf{T}$ are head pose parameters describing rotation and translation respectively and $f$ is the scale factor due to projection. We use the Swin Transformer [27] as our encoder network (Sec. 3.1). The eye-region reconstruction parameters $\mathbf{z}_M$ are used to reconstruct a textured eye-region 3D mesh, thus providing predicted 3D eyeball centres as gaze origins (eyeball vertex means), and a set of 2D projected landmarks for eye-region alignment (Sec. 3.2). The eye rotation parameters $\mathbf{z}_E$ predict the gaze vectors for both eyes. Using the gaze origins and gaze vectors, we employ a geometric vergence model to constrain the gaze directions of both eyes jointly (Sec. 3.3). Finally, we use a differentiable renderer to render the output image for pixel-wise comparison to the input. The following subsections elaborate each pipline component.

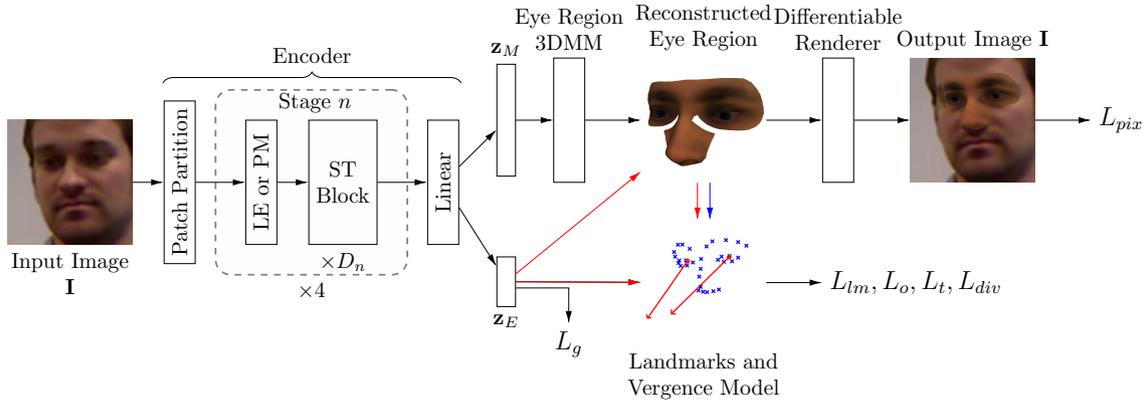

Fig. 2. Overview of gaze estimation using our active-gaze 3DMM autoencoder. We employ the tiny version of Swin Transformer for our encoder, which has four stages. *LE* stands for Linear Embedding, which is used in stage one only, and *PM*, standing for Patch Merging, is used in stages $2-4$. *ST Block* stands for Swin Transformer block. Red points are in the 3D camera coordinate system, while blue points are on the 2D image plane. The 'L' terms show where the various loss function components are generated.

### 3.1. Encoder

We employ the Swin Transformer [27] to regress eye-region features. The input image is first divided into non-overlapping patch tokens. This is followed by four Swin transformer blocks. We define $D_i$ as the number of blocks at stage $i$ and use the *Tiny* network structure provided by the authors, where $D_{1...4} = (2,2,6,2)$. For the first stage, the linear embedding module is applied before the transformer blocks, and for the other three stages, a patch merging module is applied before each set of transformer blocks to reduce the output dimensionality. These four stages jointly produce a feature map that is appended by a linear layer to regress a semantically-meaningful feature vector. This is then divided into two parts: eye-region reconstruction parameters $\mathbf{z}_M$ (Sec. 3.2) and both eyes' gaze directions $\mathbf{z}_E$ defined by azimuth and elevation (Sec. 3.3).

### 3.2. Eye-Region 3D Morphable Model (3DMM)

The eye-region 3DMM is constructed by selecting the relevant vertices and their topology from the FLAME [14] model. As shown in Fig. 3, both eyeballs, the eye-region and the nose are selected. Eyeballs are used to model gaze directions, eyeball sizes, and inter-ocular distances. The eye-region contains 22 landmarks on the eyebrows and eye contours, which is used to model eyeball positions and head poses. We omit the remaining parts of the FLAME head model, firstly to enable a more compact and efficient learning process, and secondly since they have much higher variance in features (*e.g.* mouth/jaw variations due to speech and/or facial expressions) that are not relevant to gaze modelling, and may introduce confounding factors. Notably, the largely rigid nose area, which contains nine landmarks on the nose ridge and the philtrum area, is added to strengthen the head pose prediction. We also use the texture model presented by [28, 29] to enable differentiable rendering of the eye-region model.

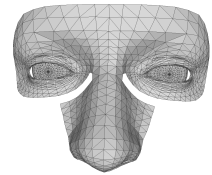

Fig. 3. The mean eye-region mesh, extracted from the FLAME model [14] and incorporated into our active-gaze 3DMM fitting system, which has rotatable eyeballs

We reconstruct the 3DMM's shape $\mathbf{S} \in \mathbb{R}^{N \times 3}$ from shape parameters $\mathbf{z}_S$ and the texture $\mathbf{A} \in \mathbb{R}^{512 \times 512 \times 3}$ from texture parameters $\mathbf{z}_A$ as follows:

$$\mathbf{S} = \boldsymbol{\mu}_S + \mathbf{U}_S \mathbf{z}_S \qquad (1)$$
$$\mathbf{A} = \boldsymbol{\mu}_A + \mathbf{U}_A \mathbf{z}_A, \qquad (2)$$

where $N$ is the number of vertices in the eye-region shape model, $\boldsymbol{\mu}_{\{S,A\}}$ and $\mathbf{U}_{\{S,A\}}$ are the mean and principal components provided by the shape and texture 3DMMs respectively. Then the eye-region shape $\mathbf{S}$ is transformed with rotation $\mathbf{R}$, translation $\mathbf{T}$ and scale $f$ to the camera coordination system by:

$$\mathbf{S}' = f\mathbf{S}\mathbf{R}^T + \mathbf{1}\mathbf{T}, \qquad (3)$$

where $\mathbf{R} \in \mathrm{SO}(3)$ is the rotation matrix derived from the rotation $r$ by Rodrigues' rotation formula and $\mathbf{1} \in \mathbb{R}^{N \times 1}$ is the vector of all ones. Finally, given the camera calibrations are available, we construct a full perspective projection $\Pi \in \mathbb{R}^3 \to \mathbb{R}^2$ to project the eye-region shape in 3D camera space $\mathbf{S}'$ to image plane, thus obtaining the predicted 2D landmarks $\hat{\mathcal{L}}$ on image plane. We use a differentiable renderer $DR$ implemented by PyTorch3D [30], with the same projection model, to form image $\hat{\mathbf{I}}$ as:

$$\hat{\mathbf{I}} = \mathrm{DR}\left(\mathbf{S}', \mathbf{A}, \Pi\right). \quad (4)$$

Note that all previous works of 3D face reconstruction that involve differentiable rendering assume a Lambertian surface, which is not well-suited to the eyeball surface due to it's inherent moisture, which causes specularities. Our further experiments shows that the geometric vergence constraints contribute the most to gaze estimation accuracy, thus we choose the ambient Phong lighting model and leave the discussion of more refined eye-region modelling in our *limitations* section.

With the reconstructed 3D eye-region, we form the 3D gaze origin loss function $L_o$ as

$$L_o = \|\hat{\mathbf{o}} - \mathbf{o}\|_1^1, \quad (5)$$

where $\mathbf{o}$ is a 3D ground truth gaze origin provided by the dataset and $\hat{\mathbf{o}}$ is some point derived by the eye-region shape. For example, a predicted eyeball centre is obtained by averaging all eyeball vertices. With such a design, our method becomes universally applicable to any gaze origin definition, as provided by the dataset; for example, both eyeball-centered and face-centered have been used in the literature. This obviates the conversion step described by Chen *et al.* [4] that converts gaze ground truth between datasets using different gaze representations.

With the projection model, 2D projected landmarks can be obtained, to give the 2D landmark loss function $L_{lm}$ as:

$$L_{lm} = \|\hat{\mathcal{L}} - \mathcal{L}\|_2^2, \quad (6)$$

where $\mathcal{L}$, the ground truth 2D landmarks, are either provided by the dataset or generated before training using PyTorch Face Landmark [31] with a pre-trained MobileNetV2 [32] as the backbone network. The predicted 2D landmarks $\hat{\mathcal{L}}$ are obtained by projecting selected vertices in the eye-region shape $\mathbf{S}'$ onto the image plane via the perspective projection, $\Pi$. We employ the Multi-PIE [33] definition of 68 face landmarks and select 31 corresponding points on the eye-region 3DMM and input images.

Finally, the pixel loss $L_{pix}$ for rendered eye-region images is formed as:

$$L_{pix} = \|\hat{\mathbf{I}} - \mathbf{I}\|_2^2. \quad (7)$$

### 3.3. Vergence Model

The predicted gaze rotations $\mathbf{z}_E = (\mathbf{r}_l, \mathbf{r}_r)^T$ by the encoder are azimuths and elevations for both eyes (*e.g.* $\mathbf{r}_l = (r_{le}, r_{la})^T$). Two rotation matrices $\mathbf{R}_{\{l,r\}}$ are derived from the rotation angles by Rodrigues' rotation formula. We assume the gaze direction is a vector originating from the centre of the eyeball, and pointing towards the iris centre. Thus, the gaze vectors for both eyes are calculated by: $\mathbf{g}_i = \mathbf{R}_i \begin{bmatrix} 0 & 0 & 1 \end{bmatrix}^T, i \in \{l, r\}$. They originate from both eyeballs' centre $\mathbf{o}_{\{l,r\}}$ respectively. The eyeball rotation matrix is also applied to the eyeball shapes of the reconstructed 3D eye-region shape to rotate the eyeballs to produce a plausible appearance.

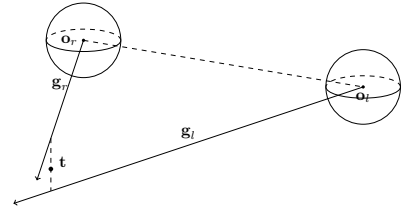

Fig. 4. The gaze-vergence model showing eyeball origins ($\mathbf{o}_{l,r}$), gaze directions ($\mathbf{g}_{l,r}$) and viewing target $\mathbf{t}$ in the global camera frame. In general, the regressed gaze directions are skew and the loss function penalises this lack of coplanarity

As shown in Fig. 4, we equip our system with geometric constraints so that both eye gazes are mutually constraining each other via a mutual gaze target $\hat{\mathbf{t}}$. Due to the nature of human gazes, there are three underlying constraints for this vergence model: i) both gaze vectors are directed away from the head; ii) the gaze vectors are coplanar; iii) the gaze vectors intersect at the gaze target $\hat{\mathbf{t}}$, unless they are parallel. These three constraints can be satisfied during the process of calculating the gaze target $\hat{\mathbf{t}}$, which is defined as the closest point between the two gaze vectors. We define $\mathbf{K}_i = \mathbf{o}_i + k_i \mathbf{g}_i, i \in \{l, r\}$ as the two end points of the shortest segment connecting left and right gazes. Therefore,

$$\hat{\mathbf{t}} = (\mathbf{K}_l + \mathbf{K}_r)/2. \quad (8)$$

Since the shortest segment must be perpendicular to both gaze vectors, we can derive the shortest distance $d$ as:

$$d := \|\mathbf{K}_l - \mathbf{K}_r\| = k_{lr}\left(\mathbf{g}_r \times \mathbf{g}_l\right), \quad (9)$$

where $k_l$, $k_r$ and $k_{lr}$ can be solved by:

$$\begin{bmatrix} k_l & k_r & k_{lr} \end{bmatrix}^T = \begin{bmatrix} \mathbf{g}_l & -\mathbf{g}_r & \mathbf{g}_r \times \mathbf{g}_l \end{bmatrix}^{-1} \left(\mathbf{o}_r - \mathbf{o}_l\right). \quad (10)$$

We design three loss terms based on the underlying constraints of the geometric vergence model. Firstly, the gaze skew loss, $L_{skew} = d^2$, encourages the two gaze vectors to be coplanar. Secondly, the predicted gaze target, $\hat{\mathbf{t}}$, along with the 3D ground truth target, $\mathbf{t}$, forms a gaze target loss $L_t = \|\hat{\mathbf{t}} - \mathbf{t}\|_1^1$. Finally, a gaze pose loss is given as $L_g = \|\mathbf{z}_E - \mathbf{r}_{gt}\|_1^1$, where $\mathbf{r}_{gt}$ is the ground truth eyeball rotation. All of these losses reduce gaze error, while preventing the physically impossible case of the gaze being directed into and behind the head.

### 3.4. Complete Loss Function

In addition to the previously stated loss function terms, we employ a regulariser on the 3D eye-region shape and texture latent code $\mathbf{z}_S$ and $\mathbf{z}_A$ to encourage the reconstructed eye-region shape and texture to stay within the model space. The regulariser is defined as follows:

$$L_{reg} = \|\mathbf{z}_S\|_2^2 + \|\mathbf{z}_A\|_2^2. \quad (11)$$

Finally, all the losses are combined linearly as $L = \Lambda^T L_{vec}$ where $\Lambda = [\lambda_1 \ldots \lambda_7]^T$ are the hyperparameter weights required to balance each loss component in the loss vector $L_{vec} = [L_{pix}, L_{lm}, L_o, L_t, L_{skew}, L_g, L_{reg}]^T$.

## 4. Evaluation

We employ the Eyediap dataset for evaluation. Eyediap [34] is a dataset containing videos of 16 subjects looking at various targets. We use the floating ball target videos where 14 distinct subjects are participating. A static head pose session and a dynamic head pose session are recorded for each subject, resulting in 28 sessions of, on average 2701 frames per session. We use the *low-resolution* version ($640 \times 480$) for our experiments. During training and testing, we utilise all validated frames except those not detected by the face landmark localisation algorithm. We perform cross-subject evaluations on this dataset, using a leave-two-subjects-out strategy by using two subjects' both static and dynamic head pose sessions as the test set, and the remainder as the training set. We train on $\sim$ 61k frames and test on $\sim$ 14k frames. The following two subsections detail our quantitative and qualitative evaluations.

The loss function we defined introduces 7 hyperparameters, which may suggests a difficult tuning process. However, many losses are imposing the same model restriction that help stabilise the training process, we group empirically correlated losses into 3 groups and treat each group as a single joint loss, to make hyperparameter tuning process simpler. We also perform ablation studies with the 3 groups in Sec. 5.

### 4.1. Quantitative Evaluation

In this section, we compare our results with some previous methods with the commonly-adopted angular error metric. This error metric measures the angle between the predicted gaze vector and the ground truth gaze vector. Results on Eyediap are given in Tab. 1 and show that our method gives the lowest mean error. We also include a baseline which uses the Swin transformer to regress gaze rotation only. Due to the difficulty of re-implementation of the appearance-based methods which employ different subjects in different video session types and over different resolutions, we cautiously show some other approaches' reported accuracy for reference.

There are two types of task for gaze vector estimation: i) the gaze originates from the eyes and ii) the gaze originates from faces [4]. Note that our method does not require explicit conversion between the eye gaze task and the face gaze task. Moreover, during training our method approaches the ground truth very quickly and we obtain our results with training for only 20 epochs on 10% of the training set (approx. 60,000 images) that is randomly sampled every batch.

Table 1. Angle error (°) on gaze vectors originate from eyeballs compared with current literature on the Eyediap dataset.

| | Method | mean ± std | median |
|---|---|---|---|
| Appearance-based Methods | Zhang et al. [26][#] | 6.76 | \ |
| | Cheng et al. [23] | 5.17 | \ |
| | Zhang et al. [35] | 7.37 | \ |
| | Sinha et al. [21] | 4.62 ± 2.93 | \ |
| | Gaze360 [36][#] | 5.58 | \ |
| | RT-Gene [37][#] | 6.30 | \ |
| | Dilated-Net [38][#] | 6.57 | \ |
| | Baseline | 5.25 ± 3.58 | 4.45 |
| Model-based Methods | PR-ALR [34][*] | 8.1 | \ |
| | Wood et al. [1][*] | 9.44 | 8.63 |
| | Ploumpis et al. [2] | 8.85 | \ |
| Combined Method | Ours | **4.55 ± 3.29** | **3.82** |

[*] Eval. on static head pose only. [#] Converted from face gaze by [4].

Lastly, we report our reconstructed model's quality. Our face patches on the Eyediap dataset have $96 \times 96$ pixels, our predicted face landmarks are filtered manually to keep frames without obstacles in front of the face. The average landmark error in pixels is 4.84 pixels per landmark. We further normalise pixel landmark errors by dividing the distance between the left eye's left corner and the right eye's right corner, which results in a proportion of 0.113.

### 4.2. Qualitative Evaluation

We present qualitative visual results on the Eyediap dataset. Four different predictions are presented in Fig. 5, each with a block of three sub figures (input image, extracted model, output image). The first shows a successful example of the static head pose. The second and the third demonstrate successful examples when different head poses are present. The final example shows a failure case where an extreme expression is present. Such expression involves massive morphing of the eye contour region. Although we have chosen the most expression-invariant parts on the face to build the eye-region model, expressions involving eyes like this are still not modelled. Note that the second and the third examples are trained with a higher weight on the pixel loss $L_{pix}$. This results in obtaining a more accurate face area texture. However, due to the nature of the albedo model we used, the eyeball's sclera region appears to be too cloudy. We consider this to be a trade-off of the albedo model and gaze estimation accuracy is not adversely affected.

## 5. Ablation Studies

Our 7 loss function components can be divided into 3 categories of task:

1. eye-region reconstruction with $L_{pix}$, $L_{lm}$ and $L_{reg}$
2. appearance-based gaze estimation with $L_g$
3. geometric vergence constraints with $L_t$, $L_o$ and $L_{skew}$

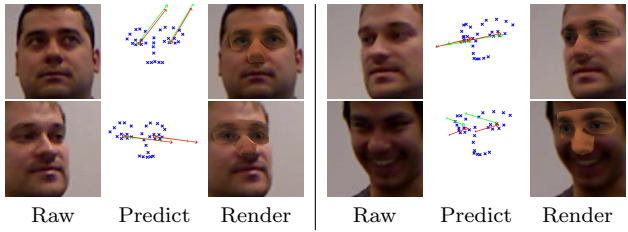

Raw   Predict   Render   Raw   Predict   Render

Fig. 5. Input images (left); model (centre) with predicted landmarks (blue crosses), predicted gaze rays (red rays), ground truth gaze rays (green rays); rendered eye-region model (right). First three subjects have successful predictions. Last subject shows a failure case with extreme expression.

We now perform ablation studies to determine the effectiveness of each task and analyse the advantages of using the state-of-the-art vision backbone network. All ablation study results are presented in Tab. 2. We used a randomly selected subject's static and dynamic head pose sessions as the test set for all ablation experiments for a fair comparison. First, we construct a baseline model that comprises of only our vision backbone network (*i.e.* Swin transformer) which predicts two eyeball rotations. It is trained with only the gaze pose loss function $L_g$, thus it solves Task 2 only. We denote this experiment as *baseline* in the table. Then we construct our system with the vergence model only, *i.e.* the model solves Task 3 only. Since the predicted gaze origins (*i.e.* eyeball centres) are not available if no 3D eye-region model is reconstructed, we predict the eyeball centres directly using the backbone network. This experiment is denoted as *Vergence model* in the table. Then we report our proposed method with all loss terms (*i.e.* aimed to solve all three tasks simultaneously), denoted as *Ours*. Lastly, we swap or remove a specific part in our proposed method to observe the impact. We remove the loss term $L_o$ to let the model learn without the ground truth eyeball positions. This experiment is denoted as *w/o $L_o$*. Finally, we evaluate the improvement in gaze estimation accuracy by employing state-of-the-art vision backbone network Swin transformer against a former popular vision backbone ResNet-18 [39]. This is denoted *Ours - ResNet18* in Tab. 2.

Table 2. Angle error (°) on subject 15 from Eyediap dataset for the ablation study

| Method | mean ± std | median |
|---|---|---|
| baseline | 5.60 ± 3.28 | 5.01 |
| Vergence model | 4.80 ± 3.07 | 4.23 |
| w/o $L_o$ | 6.64 ± 4.92 | 5.26 |
| Ours - ResNet18 | 4.94 ± 3.16 | 4.34 |
| Ours | **4.11 ± 2.93** | **3.42** |

Results in Tab. 2 shows that our proposed method, utilising all the loss components and the Swin transformer performs the best. The vanilla appearance-based method (*baseline*) cannot perform competitively when only low resolution face images are fed to the network. Our geometric vergence model contributes hugely towards an accurate gaze estimation. However, combining all three tasks performs better than using only Task 2 or Task 3, thus showing the effectiveness of our *multi-task* method.

We observe that the gaze origin loss $L_o$ is vital to the success of our proposed method. It provides the necessary guidance to both gaze direction and 3D eye-region reconstruction.

We argue that the attention mechanism of the Swin transformer [27] vision backbone network is essential to the gaze estimation task especially when only low resolution images are provided. We observe similar training times when we employ the smallest architectures for both ResNet and the Swin transformer, although the Swin transformer has 28 million parameters, while ResNet18 has 11 million.

## 6. Conclusion

Our approach reconstructs a 3D eye-region as well as the gaze direction by exploiting the advantages of both appearance-based and model-based techniques. Results show state-of-the-art gaze estimation. Our method closes the gap on the gaze estimation task where model-based methods lack the raw feature extraction ability by utilising a state-of-the-art vision backbone network. Our work can be further applied to inter-ocular distance prediction, ear-to-ear face region modelling, with highly-accurate gaze estimation. As such, our work contributes to human eye-region understanding, HCI, VR and wearables.